\documentclass{article}

\usepackage{microtype}
\usepackage{graphicx}
\usepackage{subfigure}
\usepackage{booktabs} %

\usepackage{hyperref}

\usepackage[accepted]{icml2023}

\usepackage{amsmath}
\usepackage{amssymb}
\usepackage{mathtools}
\usepackage{amsthm}

\usepackage[capitalize,noabbrev]{cleveref}

\usepackage{xcolor}

\theoremstyle{plain}

\theoremstyle{definition}

\theoremstyle{remark}

\definecolor{orange}{RGB}{255, 127, 14}
\definecolor{blue}{RGB}{31, 119, 180}
\definecolor{green}{RGB}{44, 160, 40}
\definecolor{purple}{RGB}{148, 103, 189}
\definecolor{red}{RGB}{214, 39, 40}

\usepackage[textsize=tiny]{todonotes}

\icmltitlerunning{Omega: Optimistic EMA Gradients}

\begin{document}

\twocolumn[
\icmltitle{Omega: Optimistic EMA Gradients}

\icmlsetsymbol{equal}{*}

\begin{icmlauthorlist}
\icmlauthor{Juan Ramirez}{mila,udem}
\icmlauthor{Rohan Sukumaran}{mila,udem}
\icmlauthor{Quentin Bertrand}{mila,udem}
\icmlauthor{Gauthier Gidel}{mila,udem,cifar}
\end{icmlauthorlist}

\icmlaffiliation{mila}{Mila, Québec AI Institute}
\icmlaffiliation{udem}{DIRO, Université de Montréal, Canada}
\icmlaffiliation{cifar}{Canada CIFAR AI Chair}

\icmlcorrespondingauthor{Juan Ramirez}{juan.ramirez@mila.quebec}

\icmlkeywords{Machine Learning, ICML, Min-Max Optimization, Optimistic Gradient Method, Stochastic Optimization}

\vskip 0.3in
]

\printAffiliationsAndNotice{}  %

\begin{abstract}
 Stochastic min-max optimization has gained interest in the machine learning community with the advancements in GANs and adversarial training. Although game optimization is fairly well understood in the deterministic setting, some issues persist in the stochastic regime. Recent work has shown that stochastic gradient descent-ascent methods such as the optimistic gradient are highly sensitive to noise or can fail to converge. Although alternative strategies exist, they can be prohibitively expensive. We introduce Omega, a method with optimistic-like updates that mitigates the impact of noise by incorporating an EMA of historic gradients in its update rule. We also explore a variation of this algorithm that incorporates momentum. Although we do not provide convergence guarantees, our experiments on stochastic games show that Omega outperforms the optimistic gradient method when applied to linear players. Our code is available here\footnote{Code: \href{https://github.com/juan43ramirez/Omega}{https://github.com/juan43ramirez/Omega}}.
\end{abstract}

\section{Introduction}

\label{sec:introduction}
The recent progress in machine learning can be attributed to being able to reliably minimize (or maximize) objective functions, such as losses or rewards, using gradient-based optimization. Min-max optimization has emerged as particularly relevant for machine learning due to the success of GANs~\citep{goodfellow2020generative}, adversarial training~\citep{madry2017towards}, and actor-critic systems~\citep{pfau2016connecting}. 
Even though the dynamics of gradient-based methods are well understood for minimization, some issues emerge in the context of saddle point optimization \citep{gidel2018variational}. 
Moreover, the noise introduced by using stochastic estimates of the gradients can exacerbate the issue \citep{chavdarova2019reducing}.  

Extragradient \citep{korpelevich1976extragradient} and extrapolation from the past \citep{popov1980from_the_past} are popular methods for solving deterministic min-max optimization problems as they enjoy better convergence guarantees to gradient descent-ascent \citep{gidel2018variational}. However, their natural extensions to the stochastic setting, independent samples stochastic extragradient (ISEG) \citep{gorbunov2022stochastic, beznosikov2022smooth} and independent samples optimistic gradient (ISOG) can fail to converge for a simple stochastic bilinear game and are sensitive to noise \citep{chavdarova2019reducing}. This is concerning for machine learning applications, where performing full batch optimization is prohibitively expensive.

An alternative for ISEG has been considered where the same sample is used during the extrapolation and update steps \citep{gidel2018variational, mishchenko2020revisiting}, and it has been shown to converge under weaker assumptions. For the optimistic gradient method, however, same-sample style updates would require the computation of two gradients per parameter update. Therefore, the same-sample stochastic optimistic gradient (SSOG) is equivalent to same-sample SEG in terms of its computational cost. Other methods such as variance-reduced extragradient mitigate the issues associated with stochasticity but significantly increase the cost of the algorithm~\cite{chavdarova2019reducing, alacaoglu2022stochastic}. 

\begin{figure}[h!]
    \centering
    \includegraphics[width=.22\textwidth]{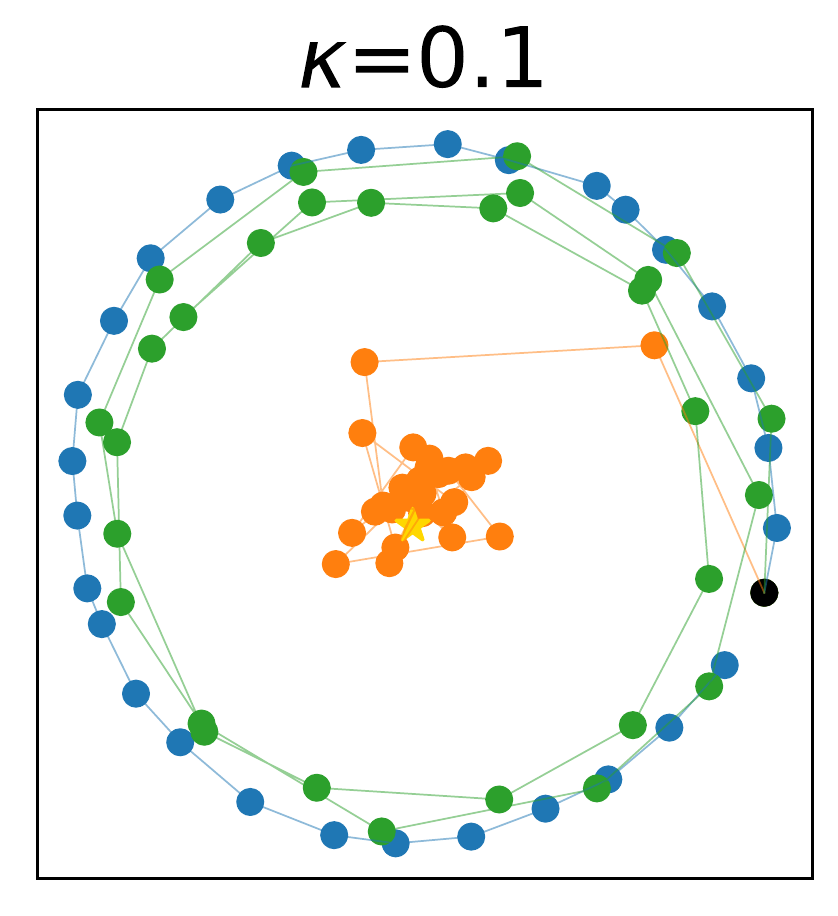}\hfill
    \includegraphics[width=.22\textwidth]{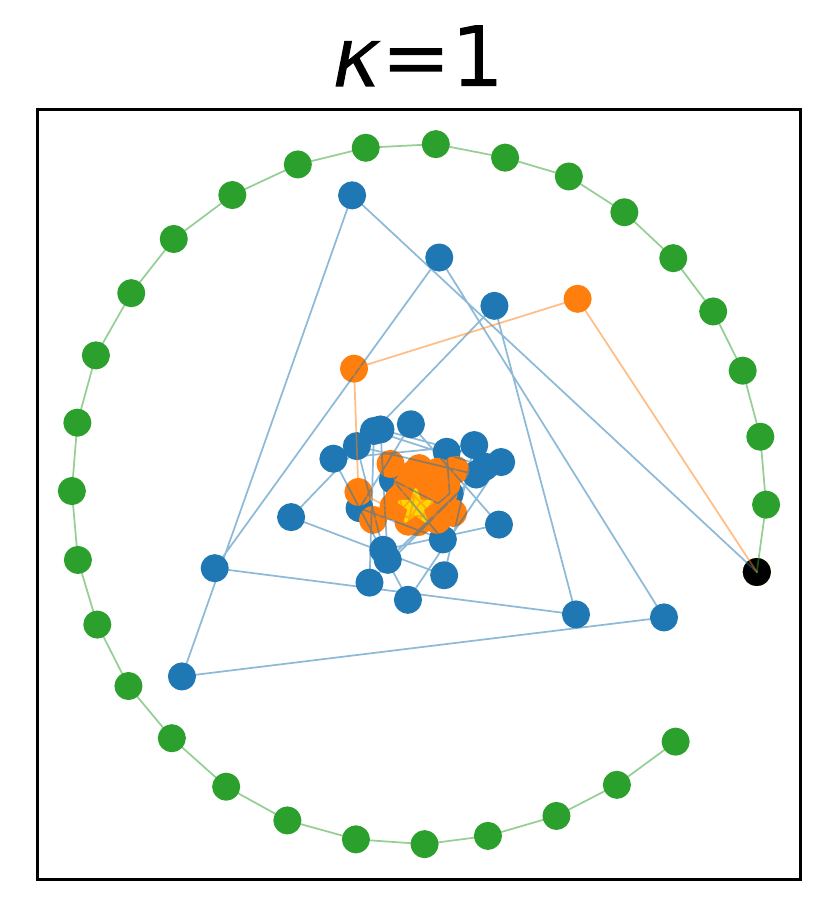}\hfill
    \caption{Iterates of stochastic gradient descent-ascent (\textcolor{green}{SGD}), the independent samples stochastic optimistic gradient method (\textcolor{blue}{ISOG}) and \textcolor{orange}{Omega} for a 2D stochastic bilinear game. $\kappa$ indicates the conditioning of the optimization problem. SGD diverges, while ISOG and Omega converge. Omega converges faster than ISOG.}  
    \label{fig:bilinear-omega-kappa}
\end{figure}

We focus on methods that employ optimistic-like updates. We propose \textit{Omega}, a variation of SOG, where an exponential moving average of historic gradients is considered in the update rule. This helps us mitigate the shortcoming of the independent samples SOG approach. More importantly, Omega requires only one gradient computation per parameter update, thereby guaranteeing better computational complexity when compared to SEG. Although we do not provide convergence guarantees for our approach, we demonstrate that Omega outperforms SOG in stochastic bilinear games (see \cref{fig:bilinear-omega-kappa}) and showcases similar performance to other methods for quadratic games. %

\textbf{Outline:} the optimistic gradient method is presented in \cref{sec:The Stochastic Optimistic Gradient Method}. Our proposed method, Omega, and its extension to incorporate momentum are introduced in \cref{sec:Optimisitc EMA gradients}. General stochastic quadratic games are described in \cref{sec:Methodology}. The performance of Omega for bilinear, quadratic and quadratic-linear games is presented in \cref{sec:exps}. Closing remarks and future directions are included in \cref{sec:conclusion}. 

\section{The Stochastic Optimistic Gradient Method}
\label{sec:The Stochastic Optimistic Gradient Method}
Consider the unconstrained stochastic saddle point optimization problem stated in \cref{min-max}, where $\xi$ terms represent the stochasticity, $x$ and $y$ represent the value of each player and $\ell$ refers to the payoff function.
\begin{equation}
    \label{min-max}
    \underset{x \in \mathbb{R}^{d_{x}}}{\text{min}}\, \, \underset{y \in \mathbb{R}^{d_{y}}}{\text{max}}\, \, \mathcal{L}(x, y) = \mathbb{E}_{\xi}\left[ \ell(x, y, \xi) \right]
\end{equation}
Let $w = [x, y]^\top$,  $F_{\xi}(w) = [\nabla_{x} \ell(x, y, \xi), - \nabla_{y} \ell(x, y, \xi)]^\top$. The update scheme of the stochastic optimistic gradient method with \textit{independent samples} (ISOG) is:
\begin{align}
    \label{eq:isog}
    w_{t+1} = w_{t} - \eta \Big[ (1 + \alpha) F_{\xi_t}(w_t) - \alpha F_{\xi_{t-1}}(w_{t-1}) \Big]
\end{align}
Where $\eta$ is a step size, $\alpha$ is the optimism hyper-parameter, and $\xi_t$ is a stochastic sample drawn at step $t$. On each step, the gradient field $F_{\xi_t}(w_t)$ is computed and stored for use on the next iteration. \Cref{eq:isog} employs independent samples $\xi_t$ and $\xi_{t-1}$ for computing the gradients at $w_t$ and $w_{t-1}$, respectively. 

The deterministic optimistic gradient method converges at a rate of the same order as extragradient \citep{korpelevich1976extragradient,gidel2018variational}, while only requiring one gradient calculation per parameter update as opposed to two \citep{gorbunov2022last}. In the stochastic setting, however, independent samples stochastic extragradient (ISEG) and ISOG require stronger assumptions for convergence, like bounded variance of the gradient field \citep{beznosikov2022smooth}. For extragradient, this assumption is unnecessary if the same stochastic sample is used for the extrapolation and update steps. The same-sample extragradient (SSEG) method does:
\begin{align}
    \label{eq:sseg}
    w_{t+1/2} &= w_{t} - \eta F_{\xi_t}(w_t) \\
    w_{t+1} &= w_{t} - \eta F_{\xi_t}(w_{t+1/2})
\end{align}
This approach computes the gradient for updating the current iterate $w_t$ at an extrapolated point $w_{t+1/2}$. The look-ahead nature of the updates in \cref{eq:sseg} is associated with stabilizing the dynamics of the min-max problem \citep{gidel2018variational}. Considering the same stochastic sample $\xi_t$ to compute the look-ahead and update steps allows SSEG to converge under weaker assumptions to ISOG \citep{beznosikov2022stochastic}. However, performing the look-ahead step requires the computation of two gradient fields per parameter update as opposed to one.

For optimistic methods, however, same-sample updates are less practical as they represent a computational overhead similar to that of extragradient. A \textit{same-sample} variant of ISOG is presented in \cref{eq:ssog}. 
Unfortunately, $F_{\xi_{t}}(w_{t-1})$ must be computed at iteration $t$ as it does not carry over from previous iterations, thus requiring two gradient calculations per parameter update. 
\begin{align}
    \label{eq:ssog}
    w_{t+1} = w_{t} - \eta \Big[ (1 + \alpha) F_{\xi_t}(w_t) - \alpha F_{\xi_{t}}(w_{t-1}) \Big]
\end{align}
Moreover, it has been noted that stochastic min-max optimization is especially sensitive to noise. \citet{chavdarova2019reducing} propose variance reduced extragradient to mitigate this issue, but their approach requires five gradient computations per parameter update on average. We are interested in an algorithm that: (i) has similar convergence properties to same-sample extragradient, (ii) is robust to noisy estimates of the gradients, and (iii) retains the computational cost of an optimistic method by requiring one gradient field calculation per parameter update.

\section{Optimistic EMA gradients}
\label{sec:Optimisitc EMA gradients}
We propose an algorithm with optimistic-like updates for stochastic min-max optimization. We consider an exponential moving average of historic gradients for the correction term in the optimistic update. This is shown in \cref{eq:omega}:
\begin{align}
    \label{eq:omega}
    w_{t+1} = w_{t} - \eta \Big[ (1 + \alpha) F_{\xi_t}(w_t) - \alpha \tilde{F}_{t-1} \Big]
\end{align}
Where $\tilde{F}_t$ is an EMA of previously observed gradients.
\begin{align}
    \label{eq:ema}
    \tilde{F}_{t} &= (1-\beta)F_{\xi_t}(w_t) + \beta \tilde{F}_{t-1}
\end{align}
A new hyper-parameter $\beta \in [0,1]$ is introduced for decaying the previous EMA value. 
We initialize $\tilde{F}_0 = F_{\xi_0}(w_0)$ so the first step of Omega corresponds to a gradient descent-ascent step. $\tilde{F}_0$ is not initialized to 0 to prevent later EMA iterates from being biased towards 0.

Note that the use of an EMA reduces the variance of the correction term $\tilde{F}_t$ when compared to $F_{\xi_t}(w_t)$. Thus, Omega is less sensitive to noisy estimates of the gradients than the optimistic gradient method. Moreover, Omega requires one gradient field calculation per parameter update, with a memory footprint associated with storing $\tilde{F}_t$. Therefore, the cost of Omega matches that of the independent samples optimistic gradient method in \cref{eq:isog}.

\subsection{Optimism and Momentum}
Alternatively, the EMA can be employed both for the update direction and the correction term. We refer to this method as optimism with momentum (or OmegaM) given that $\tilde{F}_t$ can be viewed as a momentum term with a dampening of $(1-\beta)$. The updates of this approach are:
\begin{align}
    \label{eq:omegam}
    w_{t+1} &= w_{t} - \eta\Big[ (1+\alpha) \tilde{F}_{t} - \alpha \tilde{F}_{t-1} \Big]
\end{align}
Where $\tilde{F}_0 = F_{\xi_{0}}(w_{0})$. 
It can be seen that the update rule in \cref{eq:omegam} is equivalent to \cref{eq:omega} under a specific choice of $\alpha$ (see \cref{app:omegam_is_omega}). \Cref{sec:exps} contains experiments with both Omega and Omega with momentum.
\Cref{app:adam} discusses an approach to combine Adam \citep{kingma2014adam} and optimism, as previously considered by \citet{gidel2018variational}.

\section{Stochastic Quadratic Games}
\label{sec:Methodology}

We consider the following stochastic quadratic game:
\begin{align}
    \label{eq:quad_game}
    \mathbb{E}_{\xi}\left[ \frac{1}{2}x^TA_{\xi}x + a_{\xi}x + x^TB_{\xi}y - c_{\xi}y - \frac{1}{2}y^TC_{\xi}y \right]
\end{align}
Where $x \in \mathbb{R}^{d_x}$ minimizes and $y \in \mathbb{R}^{d_y}$ maximizes.  
By ensuring that all $A_\xi$ and $C_\xi$ are symmetric positive-definite matrices, \cref{eq:quad_game} is strongly convex in $x$ and strongly concave in $y$. The bilinear coupling terms $B_\xi$ represent the interaction between both players.

In practice, we generate a fixed set of independent tuples $\{(A_i, B_i, C_i, a_i, c_i)\}_{i=1}^n$, and sample one on each iteration of the algorithm. Given $L_A \geq \mu_A > 0$ and $L_C \geq \mu_C > 0$, the $A_i$ and $C_i$ are sampled such that $\mu_{A} \preceq A_i \preceq L_{A}$ and $\mu_{C} \preceq C_i \preceq L_{C}$. The $B_i$ are generated so that their smallest singular value is $\mu_B > 0$ and their largest singular value is $L_B$. The condition numbers of each component of the game are: $\kappa_A = L_A / \mu_A$, $\, \kappa_B = L_B / \mu_B$, and $\, \kappa_C = L_C / \mu_C$. 
Our implementation of stochastic quadratic games is inspired by that of \citet{loizou2021stochastic}

\section{Experiments}
\label{sec:exps}

We solve a series of bilinear, quadratic, and quadratic-linear stochastic games. All of these are derived from \cref{eq:quad_game}. The distance to the optimal solution is measured throughout the optimization. The derivation of the optimal solution for a game is presented in \cref{app:games}. More details about the experimental setup and how the games are generated can be found in \cref{app:expts}.

\subsection{Stochastic Bilinear Game}
\label{sec:exps:bilinear}

\begin{figure}[h!]
    \vspace{-1ex}
    \centering
    \includegraphics[width=.45\textwidth]{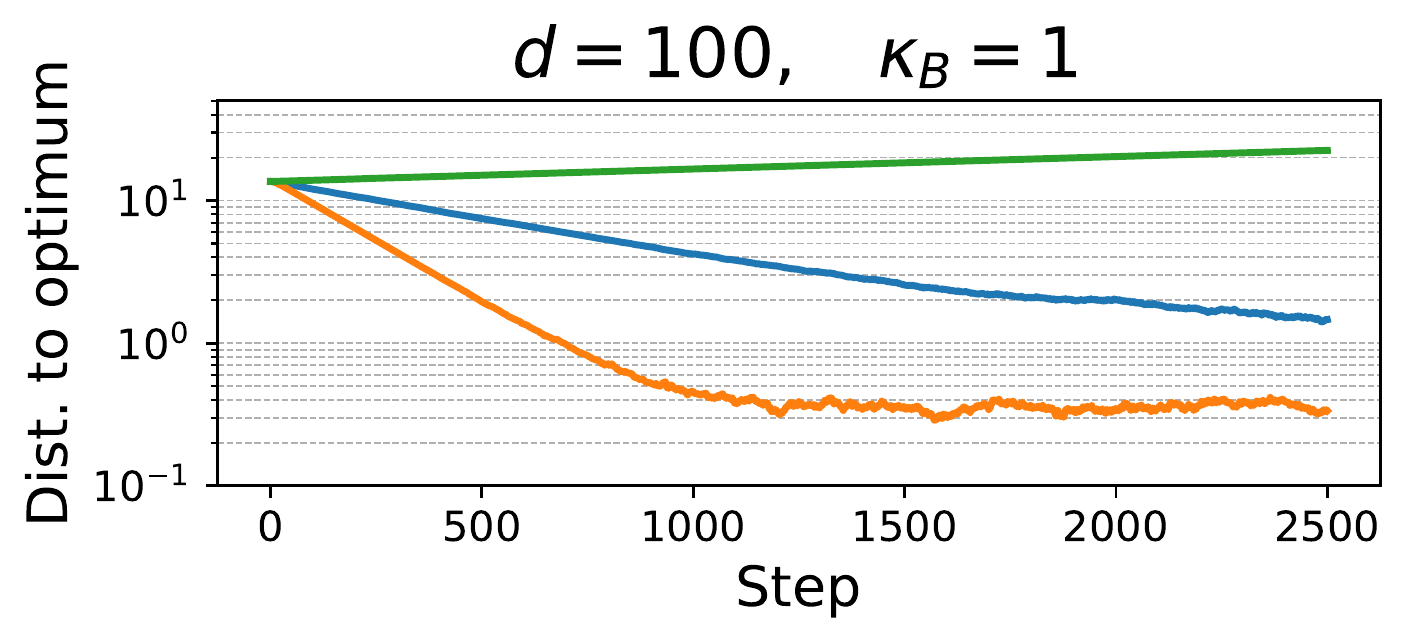}\hfill
    \caption{Distance to optimum when solving a 100-dimensional stochastic bilinear game with \textcolor{green}{SGD}, \textcolor{blue}{ISOG}, and \textcolor{orange}{Omega ($\beta=0.9$)}.\textcolor{orange}{Omega ($\beta=0.9$)} reaches the optimum faster than \textcolor{blue}{ISOG} while \textcolor{green}{SGD} diverges.  }
    \label{fig:bilinear-omega}
    \vspace{-1ex}
\end{figure}

For bilinear games, we set the terms $A$ and $C$ in \cref{eq:quad_game} to 0. \Cref{fig:bilinear-omega} shows the distance to the optimum when using SGD, ISOG, and Omega to solve a 100-dimensional bilinear game. The learning rates for ISOG and Omega are tuned separately to be 0.05 and 0.02, respectively. Although we tried OmegaM on this task, it would diverge for a wide range of learning rates.

We observe that SGD diverges while ISOG and Omega get close to the optimum and then oscillate in a neighborhood around it. Given that the optimization problem is stochastic and we use a constant step size, these oscillations are expected. In this experiment, Omega approaches the optimum faster than ISOG.

\begin{figure}[h!]
    \vspace{-1ex}
    \includegraphics[width=.45\textwidth]{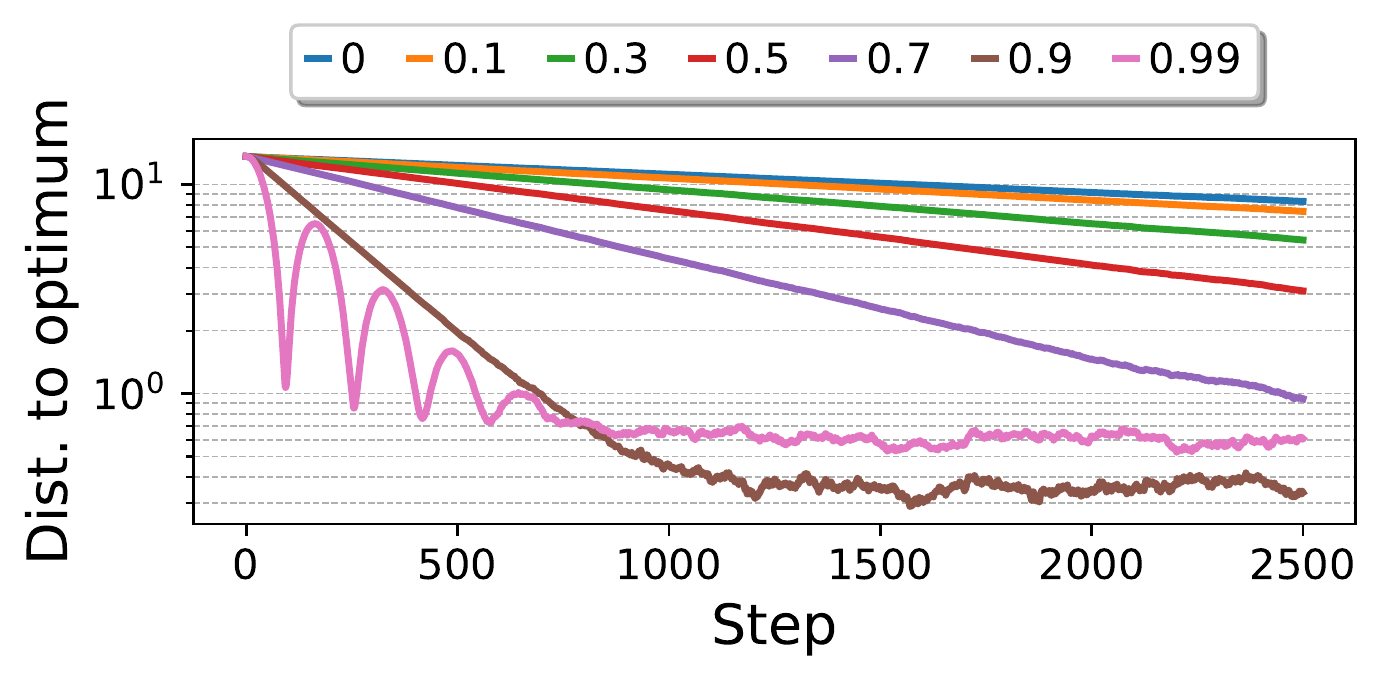}
    \centering
    \caption{Training dynamics of Omega when solving a 100-dimensional bilinear game for different choices of the EMA decay hyper-parameter $\beta$. We use a learning rate of $0.02$. We notice that an EMA of 0.99 produces larger oscillations compared to the others. An EMA of 0.9 seemed to be the best choice from the experiment, for the given setup}
    \label{fig:bilinear-ema}
    \vspace{-1.5ex}
\end{figure}

We carry out a sensitivity analysis over the EMA decay hyper-parameter $\beta$. The results are presented in \cref{fig:bilinear-ema}. Note that the experiment with $\beta=0$ corresponds to ISOG. Generally, large EMA values outperform smaller values. However, using a very large $\beta$ is detrimental as it introduces oscillations and iterates get stuck far from the optimum. Overall, we found $\beta=0.9$ to be a reasonable choice and used it throughout the rest of our work.

\Cref{tab:bilinear_batch_size} presents the distance to the optimum after 1000 steps for different choices of batch size. The learning rates for ISOG and Omega are tuned independently. Overall, both methods improve when the effect of stochasticity is mitigated by considering larger batch sizes. Notably, Omega performs well in the deterministic bilinear game. 

\begin{table}[h!]
    \centering
    \begin{tabular}{c|cllll}
    \hline
    \textbf{Batch Size} & \textbf{1}                & \multicolumn{1}{c}{\textbf{5}} & \multicolumn{1}{c}{\textbf{10}} & \multicolumn{1}{c}{\textbf{20}} & \multicolumn{1}{c}{\textbf{100}} \\ \hline
    \textbf{ISOG} & 4.224 & 3.941 & 3.876 & 3.916 & 3.900                            \\
    \textbf{Omega} & 0.458 & 0.323 & 0.291 & 0.281 & 0.280                           
    \end{tabular}
    \caption{Distance to the optimum after 1000 steps for different batch size choices. A batch size of 100 corresponds to a full batch.}
    \label{tab:bilinear_batch_size}
    \vspace{-1ex}
\end{table}

\subsection{Stochastic Quadratic Game}

\begin{figure}[!h]
    \vspace{-1ex}
    \includegraphics[width=.45\textwidth]{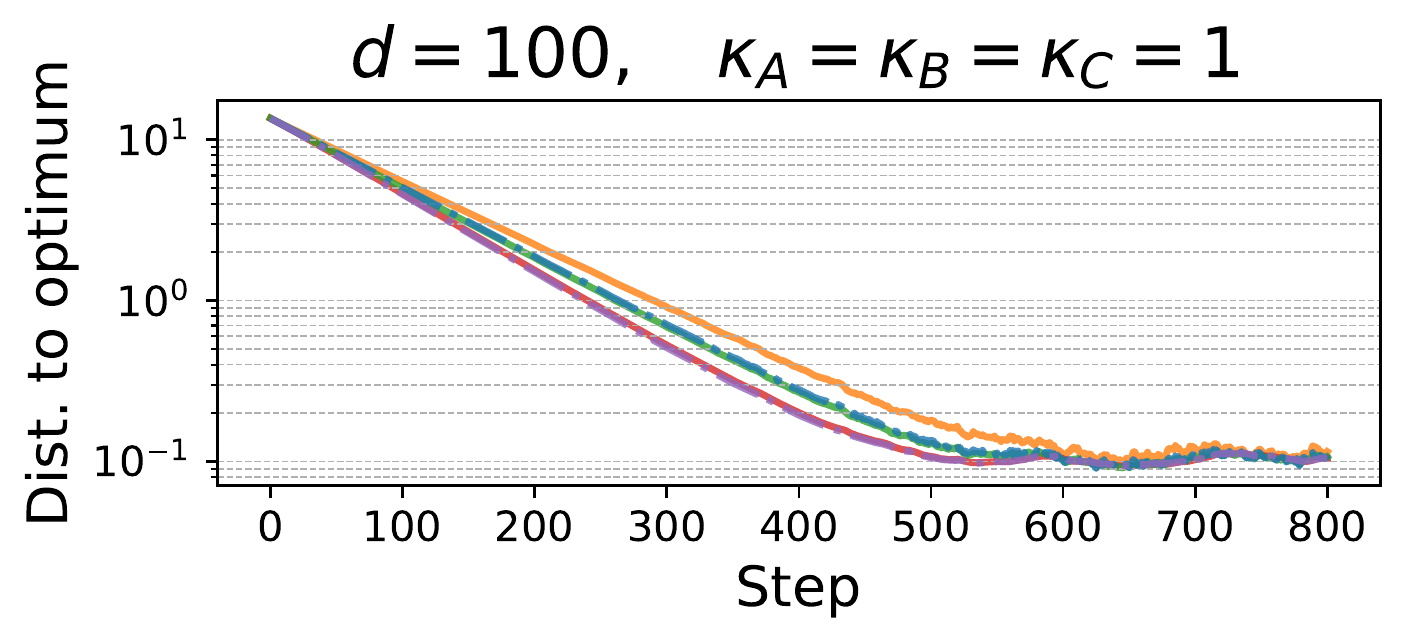}
    \centering
    \vspace{-1.5ex}
    \caption{Distance to optimum for \textcolor{green}{SGD}, \textcolor{blue}{ISOG}, \textcolor{orange}{Omega}, \textcolor{purple}{SGDM}, and \textcolor{red}{OmegaM} when solving a 100-dimensional stochastic quadratic game. For the latter 3 methods, we choose $\beta=0.9$. We can see that, given the nature of quadratic games, all algorithms converge with similar behavior.}
    \label{fig:quad}
\end{figure}

\Cref{fig:quad} shows the distance to the optimum for different optimization methods when applied to a stochastic quadratic game with good conditioning. \cref{tab:dist2opt-quad} reports the distance to the optimum after 500 steps for problems with different conditioning. We identify that a learning rate of 0.01 yields good performance for all optimization approaches.

All methods converge to a neighborhood of the optimum at a linear rate. In particular, SGDM and OmegaM yield slightly faster convergence. Although momentum can be detrimental to min-max optimization, our experiments show quadratic games as benefiting from acceleration. This could be related to the quadratic component dominating over the bilinear term and making the optimization problem easier. For quadratic games, Omega slows down progress when compared to ISOG and SGD. Similar trends are seen when considering worse conditioning for $A$ and $C$.  

\begin{table}[!h]
\centering
\resizebox{\columnwidth}{!}{%
\begin{tabular}{cclllll}
\hline
\multicolumn{1}{l}{\textbf{$\kappa_A, \, \kappa_C$}} & \textbf{$\kappa_B$} & \textbf{SGD} & \multicolumn{1}{c}{\textbf{ISOG}} & \multicolumn{1}{c}{\textbf{Omega}} & \multicolumn{1}{c}{\textbf{SGDM}} & \multicolumn{1}{c}{\textbf{OmegaM}} \\ \hline
1 &  1 & 0.119 & 0.125 & \textbf{0.112} & \textbf{0.112}  & \textbf{0.112}\\
1 & 10 & \multicolumn{1}{l}{0.128} & 0.129 & 0.135 & 0.180 & \textbf{0.106} \\
10 & 1 & 0.963 & 0.967 & \textbf{1.108} & \textbf{0.105} & 0.116 \\
10 & 10 & \multicolumn{1}{l}{0.915} & 0.908 & 1.035 & \textbf{0.777} &  0.820                            
\end{tabular}
}
\caption{Distance to the optimum at 500 steps for stochastic quadratic games with different conditioning.}
\label{tab:dist2opt-quad}
\vspace{-1ex}
\end{table}

\subsection{Quadratic-Linear Games}

This section considers a game that is quadratic on one player, but linear on the other player. Min-max optimization with one linear player is relevant for the Lagrangian-based constrained optimization literature \citep{elenter2022lagrangian}, as the Lagrangian is always a linear function of the dual variables.  
Given that Omega performs well in bilinear games but under-performs in quadratic games, we fix SGD with a step size of 0.02 for the quadratic player and experiment with the choice of optimizer for the linear player. 
 
\begin{figure}[h!]
    \vspace{-1ex}
    \centering
    \includegraphics[width=.45\textwidth]{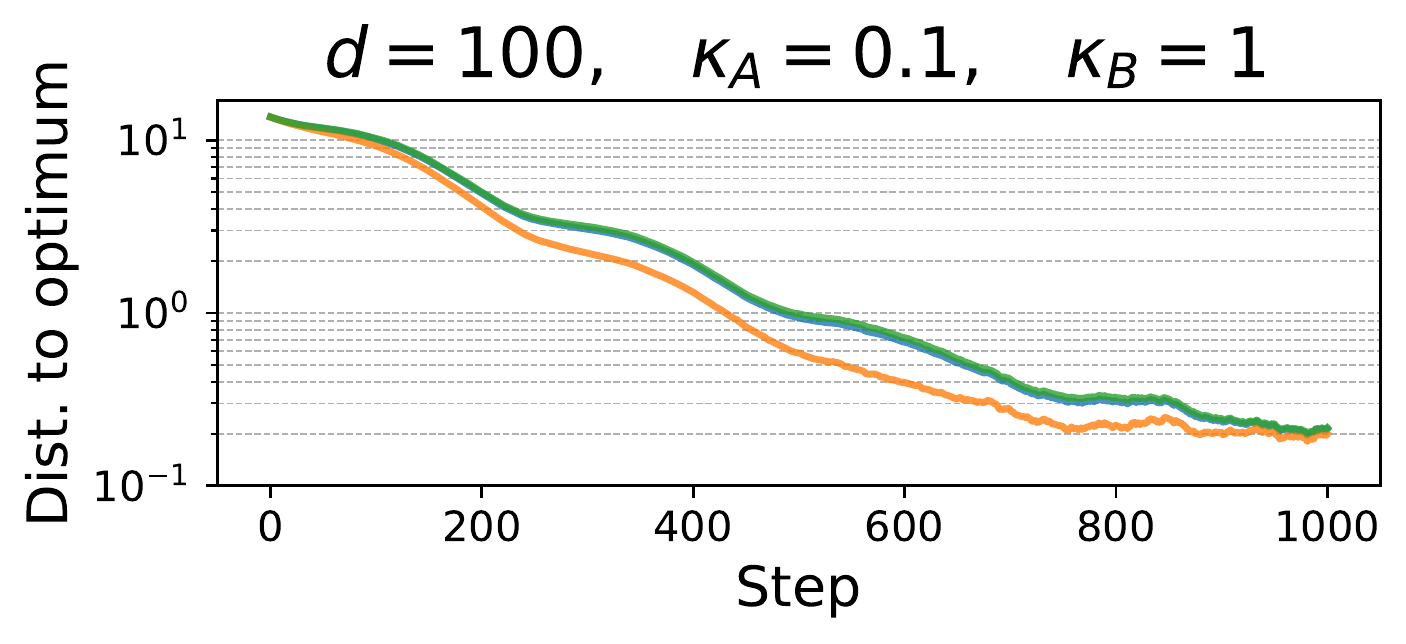}\hfill
    \vspace{-1.5ex}
    \caption{Distance to optimum when solving a 100-dimensional stochastic quadratic-linear game. SGD is used for the quadratic player, and \textcolor{green}{SGD}, \textcolor{blue}{ISOG}, and \textcolor{orange}{Omega} for the linear player. We can notice that Omega makes progress toward the optimum faster than the other methods.}
    \label{fig:quadratic-linear}
\end{figure}

\Cref{fig:quadratic-linear} shows the distance to the optimum when solving a quadratic-linear game and using SGD, ISOG, and Omega for the linear player, all with a step size of 0.01.
While SGD and ISOG perform very similarly, the Omega experiment approaches the optimum at a slightly faster rate.

\Cref{tab:dist2opt-quad-lin} reports the distance to the optimum after 500 steps for problems with different conditioning. Omega generally approaches the optimum faster than SGD and ISOG. This is most notorious when the conditioning of the quadratic term is relatively poor ($\kappa_A=10$, $\kappa_B=1$). 

\begin{table}[!h]
\centering
\resizebox{0.6\columnwidth}{!}{%
\begin{tabular}{cclll}
\hline
\multicolumn{1}{l}{\textbf{$\kappa_A$}} & \textbf{$\kappa_B$} & \textbf{SGD} & \multicolumn{1}{c}{\textbf{ISOG}} & \multicolumn{1}{c}{\textbf{Omega}} \\ \hline
1 &  1 & 0.225 & 0.221 & 0.176 \\
1 & 10 & 2.691 & 2.689 & 2.664 \\
10 & 1 & 0.990 & 0.941 & 0.587 \\
10 & 10 & 1.247 & 1.242 & 1.198                            
\end{tabular}
}
\caption{Distance to the optimum at 500 steps for stochastic quadratic-linear games with different conditioning.}
\vspace{-1ex}
\label{tab:dist2opt-quad-lin}
\end{table}

\section{Conclusion}
\label{sec:conclusion}
We consider algorithms for stochastic min-max optimization. Given that extragradient methods are computationally expensive, we focus on updates based on optimism. However, as independent samples SOG can fail to converge in some settings, and same-sample SOG is as expensive as stochastic extragradient, we propose Omega, a simple modification of the optimistic gradient method. By considering an EMA of past gradients for the update, we obtain an algorithm that has the computational requirements of an optimistic method and that is competitive on a set of toy experiments. In particular, Omega outperforms the optimistic gradient method when applied to linear players in stochastic games.

An important step for future work is to analyze the convergence properties of Omega. 
Moreover, it would interesting to evaluate Omega on a machine learning task such as training of GANs \citep{goodfellow2020generative}. 

\section*{Acknowledgements}
We would like to thank Golnoosh Farnadi for her suggestions for the writing of the paper. This research was enabled in part by compute resources, software, and technical help provided by Mila (mila.quebec). We would also like to thank the anonymous reviewers of the LXAI workshop for their constructive comments.

\bibliography{example_paper}
\bibliographystyle{icml2023}

\newpage
\appendix
\onecolumn
\section{Optimistic-like Updates.}

\subsection{Omega with Momentum is Equivalent to Omega}
\label{app:omegam_is_omega}
It can be seen that the update rule in \cref{eq:omegam} is equivalent to \cref{eq:omega} under a specific choice of the optimism hyper-parameter $\alpha$. By replacing $\tilde{F}_{t}$ with its definition, given in \cref{eq:ema}, it follows that:
\begin{align}
    w_{t+1} &= w_{t} - \eta\Big[ \tilde{F}_{t} + \alpha \big( \tilde{F}_{t} - \tilde{F}_{t-1} \big) \Big] \\
    &= w_{t} - \eta\Big[ (1-\beta) F_{\xi_t}(w_t) + \beta \tilde{F}_{t-1} + \alpha \big[ (1-\beta) F_{\xi_t}(w_t) + \beta \tilde{F}_{t-1} - \tilde{F}_{t-1} \big] \Big] \\
    &= w_{t} - \eta\Big[ (1-\beta) F_{\xi_t}(w_t) + \beta \tilde{F}_{t-1} + \alpha (1-\beta) \big[ F_{\xi_t}(w_t) - \tilde{F}_{t-1} \big] \Big] \\
    &= w_{t} - \eta\Big[ F_{\xi_t}(w_t) - \beta \big[F_{\xi_t}(w_t) - \tilde{F}_{t-1}  \big] + \alpha (1-\beta) \big[ F_{\xi_t}(w_t) - \tilde{F}_{t-1} \big] \Big]\\
    &= w_{t} - \eta\Big[ F_{\xi_t}(w_t) + (\alpha (1-\beta) - \beta) \big[F_{\xi_t}(w_t) - \tilde{F}_{t-1}  \big]  \Big] 
\end{align}
Which is equivalent to Omega with a choice of optimism of $\alpha (1-\beta) - \beta$ and with the same EMA coefficient. 

\subsection{Optimism and Adam}
\label{app:adam}

Adam \citep{kingma2014adam} is a popular method for stochastic minimization that has been shown to work well for min-max problems such as training Generative Adversarial Networks \citep{goodfellow2020generative}. Adam updates can be written as:
\begin{align}
    \tilde{F}_t &= (1 - \beta)F_{\xi_t}(w_t) + \beta \tilde{F}_{t-1} \\ 
    v_t &= (1 - \gamma) F_{\xi_t}(w_t)^2 + \gamma v_{t-1} \\
    \label{eq:adam}
    w_{t} &= w_{t- 1} - \eta \, \frac{\tilde{F}_t}{\sqrt{v_t} + \epsilon}
\end{align}
Where $v_t$ is an EMA on the square of the gradients with coefficient $\gamma \in [0,1]$, and $\epsilon > 0$ is a small number.
The usual Adam updates apply a correction to the first and second order estimates to account for their bias towards 0. 
\begin{align}
    \hat{F_t} = \tilde{F}_t / (1 - \beta ^t) \quad \hat{v_t} = v_t / (1 - \gamma ^t)
\end{align}
We disregard this correction in our presentation of the updates in \cref{eq:adam} as we choose to initialize $\tilde{F}_0 = F_{\xi_{0}}(w_{0})$ and  $v_0 = F_{\xi_{0}}(w_{0})^2$, as opposed to starting both at 0. 

Combining optimistic-like updates with Adam yields the following update rule:
\begin{align}
    \label{eq:optimistic_adam}
    w_{t} &= w_{t- 1} - \eta \left[ \, \frac{\tilde{F}_t}{\sqrt{v_t} + \epsilon}  + \alpha \left( \frac{\tilde{F}_t}{\sqrt{v_t} + \epsilon} - \frac{\tilde{F}_{t-1}}{\sqrt{v_{t-1}} + \epsilon} \right) \right]
\end{align}

This formulation already considers EMAs on the gradients. As such, optimistic Adam does not need to be extended to Omega-style updates. Note that the pairing of optimistic updates with Adam presented in \cref{eq:optimistic_adam} has been considered in the past by \cite{gidel2018variational}.

\section{Stochastic Games}
\label{app:games}

\subsection{Solving the Stochastic Quadratic Game}
\label{app:quad_games}

The stochastic quadratic game is given by:
\begin{align}
    \label{eq:quad_appx}
    \underset{x \in \mathbb{R}^{d_{x}}}{\text{min}}\, \, \underset{y \in \mathbb{R}^{d_{y}}}{\text{max}}\, \, 
 \mathbb{E}_{\xi}\left[ \frac{1}{2}x^TA_{\xi}x + a_{\xi}x + x^TB_{\xi}y - c_{\xi}y - \frac{1}{2}y^TC_{\xi}y \right]
\end{align}
\cref{eq:quad_appx} represents a strongly convex-strongly concave game if $\mathbb{E}_\xi [A_\xi]$ and
$\mathbb{E}_\xi [C_\xi]$ are positive definite matrices with their smallest eigenvalues being $\mu_A>0$ and $\mu_C>0$, respectively. In such a setting, the min-max optimization problem has a unique Nash equilibrium. Given a dataset $\{(A_i, B_i, C_i, a_i, c_i)\}_{i=1}^n$, the Nash equilibrium can be derived analytically by considering:
\begin{align}
    &\bar{A} = \frac{1}{n} \sum_{i=1}^n A_i \qquad \bar{B} = \frac{1}{n} \sum_{i=1}^n B_i \qquad \bar{C} = \frac{1}{n} \sum_{i=1}^n C_i \\ & \quad \qquad \qquad \bar{a} = \frac{1}{n} \sum_{i=1}^n a_i \qquad \bar{c} = \frac{1}{n} \sum_{i=1}^n c_i
\end{align}
And finding a stationary point of \cref{eq:quad_appx} by solving the following linear system:
\begin{align}
    \begin{bmatrix}
        \bar{A} & \bar{B} \\
        -\bar{B}^\top & \bar{C}
    \end{bmatrix}
    \begin{bmatrix}
        x \\ y
    \end{bmatrix} +
    \begin{bmatrix}
        \bar{a} \\
        \bar{c}
    \end{bmatrix} =
    0
\end{align}

\subsection{Solving the Stochastic Bilinear Game}
\label{app:bilinear_games}

The stochastic bilinear game is given by:
\begin{align}
    \label{eq:bilinear}
    \underset{x \in \mathbb{R}^{d_{x}}}{\text{min}}\, \, \underset{y \in \mathbb{R}^{d_{y}}}{\text{max}}\, \, 
 \mathbb{E}_{\xi}\left[a_{\xi}x + x^TB_{\xi}y - c_{\xi}y \right]
\end{align}
Consider a dataset $\{(B_i, a_i, c_i)\}_{i=1}^n$. Moreover, let:
\begin{align}
    \bar{a} = \frac{1}{n} \sum_{i=1}^n a_i \qquad \bar{B} = \frac{1}{n} \sum_{i=1}^n B_i \qquad \bar{c} = \frac{1}{n} \sum_{i=1}^n c_i
\end{align}
The first order necessary conditions for the problem in \cref{eq:bilinear} are $\bar{B}y + \bar{a} = 0 \text{ and } x^\top \bar{B} + \bar{c} = 0$. If, in addition, $\bar{B}$ has full rank, each system of equations has a unique solution.
\section*{Contribution of LatinX Individuals}
The first author of this work is a LatinX individual.

\section{Experimental Details}
\label{app:expts}
We use Pytorch 1.13 \citep{pytorch}. Our implementation of the optimistic gradient method is a slight modification of that provided by \citet{gidel2018variational}\footnote{Optimistic gradient method code available here: \href{https://github.com/GauthierGidel/Variational-Inequality-GAN/blob/master/optim/omd.py}{https://github.com/GauthierGidel/Variational-Inequality-GAN}}. Our code for Omega is inspired by their implementation of optimism.

\subsection{Games}
We perform experiments on bilinear, quadratic and quadratic-linear games. The default hyper-parameters for these problems are presented in \cref{tab:games}. We focus on games where both player vectors have 100 dimensions. We consider 100 stochastic samples for the matrices and vectors of the game. 

\begin{table}[h]
\centering
\begin{tabular}{|c|ccc|cc|cc|cc|}
\hline
\textbf{Game} & $d_x$ & $d_y$ & $n$ & $\mu_A$ & $L_A$ & $\mu_B$ & $L_B$ & $\mu_C$ & $L_C$ \\
\hline
\textbf{Bilinear} & 100 & 100 & 100 & - & - & 1 & 1 & - & -\\
\textbf{Quadratic} & 100 & 100 & 100 & 1 & 1 & 1 & 1 & 1 & 1 \\
\textbf{Quadratic-Linear} & 100 & 100 & 100 & 1 & 1 & 1 & 1 & - & - \\
\hline
\end{tabular}
\caption{Default game hyper-parameters.}
\label{tab:games}
\end{table}

\subsection{Generating the Stochastic Games}

We generate a fixed set of independent tuples $\{(A_i, B_i, C_i, a_i, c_i)\}_{i=1}^n$ to sample from during each iteration of the algorithm. Given $L_A \geq \mu_A > 0$ and $L_C \geq \mu_C > 0$, we generate $A_i$ and $C_i$ such that:
\begin{align}
    \mu_{A} \cdot \mathbb{I}_{d_x} \preceq &\, A_i \preceq L_{A}\cdot \mathbb{I}_{d_x} \\ \mu_{C}\cdot\mathbb{I}_{d_y} \preceq &\, C_i \preceq L_{C}\cdot\mathbb{I}_{d_y}
\end{align}
The $B_i$ are generated so that their smallest singular value is $\mu_B > 0$ and their largest singular value is $L_B$. We consider the following condition numbers throughout our work: 
\begin{align}
    \kappa_A = L_A / \mu_A, \quad \kappa_B = L_B / \mu_B, \quad \kappa_C = L_C / \mu_C
\end{align}
For bilinear games, we set $A_i=C_i=0$. Each entry in $a_i$ is drawn independently from a Gaussian distribution with mean 0 and variance $1/d_x$. The $c_i$ terms are drawn analogously.   

The initial guess for the algorithm $(x_0,y_0)$ is initialized from an isotropic Gaussian distribution with unit variance.

\subsection{Optimization Hyper-Parameters}
Different optimization algorithms are considered for solving stochastic games: gradient descent-ascent (SGD), gradient descent-ascent with momentum (SGDA, see \cref{app:sgdm}), the independent samples optimistic gradient method (ISOG) from \cref{eq:isog}, Omega, and Omega with momentum from \cref{eq:omegam}. We refer to Omega with momentum as \textit{OmegaM}. 

For ISOG, Omega, and OmegaM, we set the optimism hyper-parameter to 1. For the approaches with EMAs, namely, SGDM, Omega, and OmegaM, we employ a decay hyper-parameter $\beta$ of $0.9$.
For all optimization methods, we employ simultaneous updates. Experiments are run for 5000 steps with a batch size of 1. We only consider one random seed per experiment. Although a constant step size for stochastic optimization allows for convergence up to a neighborhood of the optimal point \citep{schmidt2014convergence}, our experiments do not employ decreasing step sizes.

\subsection{Our Implementation of SGD with Momentum}
\label{app:sgdm}
We consider an implementation of SGD with momentum which incorporates dampening:
\begin{align}
    \label{eq:sgdm}
    d_t &= \beta d_{t-1} + (1 - \tau) F_{\xi_{t-1}}(w_{t-1}) \\
    w_{t+1} &= w_t - \eta d_t
\end{align}
Where $\beta$ is the momentum coefficient and $\tau$ is the dampening. By letting $\beta \in [0,1]$ and $\tau = \beta$, we recover that updates are performed in the direction of an EMA of historic gradients: 
\begin{align}
    \label{eq:sgd_ema}   
    \tilde{F}_t &= \beta \tilde{F}_{t-1} + (1 - \beta)F_{\xi_t}(w_t) \nonumber \\ 
    w_{t+1} &= w_t - \eta \tilde{F}_t
\end{align}
For ease of comparison with Omega and Omega with momentum, our experiments with SGD with momentum consider the formulation presented in \cref{eq:sgd_ema}.

\end{document}